\pdfoutput=1

\documentclass[11pt]{article}

\usepackage[]{ACL2023}

\usepackage{times}
\usepackage{latexsym}
\usepackage{amsmath}

\usepackage[T1]{fontenc}

\usepackage[utf8]{inputenc}

\usepackage{microtype}

\usepackage{inconsolata}
\usepackage{booktabs}
\usepackage{enumitem}
\usepackage{lipsum}
\usepackage{amssymb}
\usepackage{graphicx}
\usepackage{adjustbox} 
\usepackage{array}
\usepackage{multirow}

\newcommand*\RETRO{\textsc{Retro}}

\title{Studying the Role of Input-Neighbor Overlap in \\ Retrieval-Augmented Language Models Training Efficiency}

\author{Ehsan Doostmohammadi \and Marco Kuhlmann \\
        Linköping University, Sweden \\
        \href{mailto:ehsan.doostmohammadi@liu.se}{\texttt{ehsan.doostmohammadi@liu.se}}}

\begin{document}

\maketitle

\begin{abstract}
Retrieval-augmented language models have demonstrated performance comparable to much larger models while requiring fewer computational resources. The effectiveness of these models crucially depends on the overlap between query and retrieved context, but the optimal degree of this overlap remains unexplored. In this paper, we systematically investigate how varying levels of query--context overlap affect model performance during both training and inference. Our experiments reveal that increased overlap initially has minimal effect, but substantially improves test-time perplexity and accelerates model learning above a critical threshold. Building on these findings, we demonstrate that deliberately increasing overlap through synthetic context can enhance data efficiency and reduce training time by approximately 40\% without compromising performance. We specifically generate synthetic context through paraphrasing queries. We validate our perplexity-based findings on question-answering tasks, confirming that the benefits of retrieval-augmented language modeling extend to practical applications. Our results provide empirical evidence of significant optimization potential for retrieval mechanisms in language model pretraining.
\end{abstract}

\section{Introduction}

Language models that are pretrained with retrieval augmentation can match the performance of much larger models trained in the conventional way, while at the same time requiring significantly fewer computational resources \cite{borgeaud2022improving,izacard2023atlas}.
In retrieval-augmented pretraining, the model can query and incorporate information from external sources, which makes it easier to update its knowledge base and allows information to be added, removed, or modified in a transparent and flexible way \cite{izacard2023atlas, wang2023survey, 10.1145/3627673.3679722}.

While research on retrieval-augmented language models has shown that accessing external sources reduces reliance on model parameters and leads to lower perplexity, questions remain about the underlying mechanisms driving these improvements.
Recent work has explored this issue with a focus on the role of the retrieved context \cite{borgeaud2022improving,norlund-etal-2023-generalization,doostmohammadi-etal-2023-surface}.
Findings suggest that the primary reason for reduced perplexity is \emph{surface-level overlap}, i.e., exact token matches, between the queries and the retrieved context. Yet, the optimal degree of overlap is still unclear.
Intuitively, while higher overlap appears to provide a stronger signal for language modeling, excessive similarity between queries and retrieved context may lead to over-reliance on retrieval and reduce model generalization in downstream tasks.
This raises a fundamental question: what makes retrieved context effective during pretraining?
An answer to this question could open the door to a well-founded methodology for designing retrieval corpora to maximize their usefulness for practical applications and training retrieval-augmented systems that rival the performance of much larger conventional language models under significantly tighter resource constraints---making advanced capabilities more accessible, adaptable, and sustainable.

In this paper, we take a significant step towards a deeper understanding of the role of retrieved context in augmented language modeling by systematically exploring how the degree of overlap between queries and context affects model performance both at training and at test time.
To this end, we train multiple models under controlled levels of overlap and evaluate them in terms of perplexity and on downstream tasks.
Building on our findings, we further investigate to what extent we can deliberately accelerate learning and enhance model performance in a low-resource scenario through data synthesis.

\paragraph{Contributions}

Our contributions are as follows:

\begin{itemize}[leftmargin=*]
    \item We investigate how varying degrees of overlap between queries and retrieved context affect test-time perplexity. Additionally, we analyze this variation over training steps, offering insights into how the impact of overlap depends on the amount of training data.

    \item To validate our findings, we include downstream performance results on a question answering task \cite{kwiatkowski-etal-2019-natural}, ensuring that the observed trends translate to real-world utility.

    \item We finally show how our findings can be used to train retrieval-augmented language models more data-efficiently than standard models. Specifically, we explore a method where we deliberately increase query--context overlap using synthetic contexts obtained through paraphrasing and find that this leads to faster perplexity reduction with less data.
\end{itemize}

\section{Previous Work}

Retrieval augmentation has been widely used in open-domain question answering and has also been applied to the pretraining and finetuning of language models \cite{karpukhin-etal-2020-dense, yogatama-etal-2021-adaptive, borgeaud2022improving, izacard2023atlas, wang2023shall, shi2024incontext}.

Early work, such as that of \citet{10.5555/3524938.3525306}, explored retrieve-and-edit paradigms, while follow-up studies focused on selecting relevant evidence based on lexical overlap \cite{Asai2020Learning} or enhancing inference-time generation with retrieval \cite{Khandelwal2020Generalization, 10.1162/tacl_a_00371}.
One line of work, exemplified by kNN-LM \cite{Khandelwal2020Generalization}, interpolates between model predictions and retrieved contexts at generation time.
This approach was later extended in SPALM \cite{10.1162/tacl_a_00371}, which introduced a learned gating mechanism that dynamically balances between both contributions.

Later efforts have shifted toward integrating retrieval earlier in the training pipeline. \citet{borgeaud2022improving} demonstrated that large-scale retrieval-augmented pretraining can substantially reduce perplexity even with a frozen retriever.
\citet{izacard2023atlas} further showed that jointly training the retriever and language model can provide additional performance gains, especially when retrieval is over extremely large datasets (trillions of tokens).
\citet{10.5555/3618408.3620004} found that approximate nearest neighbors have a positive effect on generalization, acting as a form of regularization.

Recent work has found that retrieval-augmented pretraining leads language models to acquire less world knowledge but improved syntactic proficiency \cite{samuel-etal-2024-room}.
This shows that such training shifts the role of the language model toward interpreting factual information from retrieved contexts, which, in practice, offloads knowledge from the model parameters and allows the use of smaller model sizes.
Although the majority of published retrieval-augmented systems rely on relatively small language models, they still demonstrate significant improvements in perplexity, factuality, and downstream accuracy when pretraining is retrieval-enhanced \cite{borgeaud2022improving, izacard2023atlas}.
These findings suggest that retrieval-augmented pretraining is a promising direction for scaling language models more efficiently than approaches based on parameters alone.


\section{Background: \RETRO{} Architecture}

In this paper, we experiment with retrieval-augmented language models based on the \RETRO\ architecture \citep{borgeaud2022improving}.
This architecture is similar to GPT but is set up to predict the next token conditioned on an augmented context that, in addition to the previously generated tokens, includes additional tokens obtained via the retrieval mechanism.
Technically, this is implemented via an additional cross-attention mechanism between the internal representations of the generated tokens and the encoded context.
This design allows the model to incorporate information from the retrieval database without requiring it to be explicitly included in the generated token sequence.

\paragraph{Chunks}

The retrieval of additional context and its incorporation into next-token prediction is done at the level of \emph{chunks}.
A chunk is defined as a contiguous sequence of tokens with a fixed size, which is set as a hyperparameter.
In both the original \RETRO\ paper \citep{borgeaud2022improving} and our own work, the chunk size is $m = 64$.

\paragraph{Neighbors}

When the model has generated a new chunk $C_u$, that chunk is used as a query to retrieve $k$ similar chunks from the retrieval database.
In this context, the chunk $C_u$ is conventionally called the \emph{input chunk}, and the retrieved chunks are called the \emph{neighbors} of~$C_u$.
Each neighbor $N_u^i$ is additionally concatenated with the chunk $F_u^i$ that follows $N_u^i$ in the retrieval dataset; that chunk is called the \emph{continuation} of the neighbor.
The rationale of the augmentation is that, since the neighbors are retrieved based on their similarity to $C_u$, their continuations are likely to be similar to $C_{u+1}$, the next chunk to be generated by the model, and should therefore be able to inform the generation of that chunk.
For convenience, we generally use the term \emph{neighbor} to include both the neighbor proper and its continuation.

\paragraph{Retrieval-augmented context}

In the following, we write $\text{\textsc{Ret}}(C_u)$ to denote the retrieval-augmented context that the model uses to generate the tokens in the chunk $C_{u+1}$.
The generation of the first chunk $C_1$ is not conditioned on any augmented context (only on the usual language modeling context), so $\text{\textsc{Ret}}(C_0) = \emptyset$.
For $u \geq 1$, the retrieval-augmented context is
\begin{displaymath}
    \text{\textsc{Ret}}(C_u) \triangleq ([N_u^1, F_u^1], \ldots, [N_u^k, F_u^k])\,.
\end{displaymath}
Note that in \RETRO, the retrieval happens off-line and does not involve any trainable parameters within the model, unlike some other approaches such as \textsc{Atlas} \cite{izacard2023atlas}.

\section{Experimental Framework}

In this section, we describe the components of our experimental framework that are shared across all of our experiments.

\subsection{\RETRO-fitting}

While we could train retrieval-augmented language models from scratch, here we instead opt to train models by continued pretraining of a GPT-style base model with \RETRO-style augmented retrieval.
We refer to this process as \emph{\RETRO-fitting}.
With the rise of strong open-source foundation models, \RETRO-fitting is more realistic for real-world applications than full pretraining.
It also enables us to conduct more experiments, as it significantly reduces training time.
Moreover, \citet{borgeaud2022improving} show that \RETRO-fitted models can achieve a perplexity and downstream performance that is comparable to that achieved with full training.

Technically, \RETRO-fitting entails expanding the base model with two new types of layers:
\begin{enumerate}[leftmargin=*, itemsep=0pt]
\item an encoder for the retrieved context chunks (neighbors and their continuations); and
\item cross-attention between the retrieved context and the standard language modeling context.
\end{enumerate}
These layers are randomly initialized and trained alongside the rest of the GPT-initialized weights.

\subsection{Models}
\label{sec:FrameworkModels}

For all our experiments, we \RETRO-fit a 345M parameter GPT model pretrained by Nvidia \cite{shoeybi2019megatron}.
This model has 24 transformer layers, each with a hidden size of 1,024 and 16 attention heads, similar to the GPT-2 medium model \cite{radford2019language}.
We chose to go with a relatively small base model because we want to specifically explore the potential of offloading information to the retrieval mechanism rather than storing it in the model parameters, which is a core motivation behind retrieval-augmented pretraining.
Indeed, previous work on downstream question answering tasks has shown that \RETRO\ sees the more benefit from retrieval the fewer parameters it has to store information in \citep{wang2024instructretro}.
Also, small models allow us to do more experiments given a fixed computational budget, and \RETRO\ shows similar perplexity curves regardless of model size \cite{borgeaud2022improving}.

\subsection{Retrieval}

For retrieval, we use the training set of the Pile \cite{gao2020pile}, which comprises about 800~GB of text of different genres.
We embed this data using mean pooling over representations from \texttt{MiniLM-L6-H384-uncased} \citep{wang2020minilm}.
This model is one of the top performers at sentence embedding according to measurements on Sentence Transformers \cite{reimers-2019-sentence-bert}.
To perform indexing and approximate search, we use FAISS \cite{johnson2019billion} with the index configuration \texttt{OPQ32\_64}, \texttt{IVF65536\_HNSW8}, \texttt{PQ32} to enable efficient and scalable approximate nearest neighbor search.
This configuration first applies Optimized Product Quantization (OPQ) to rotate and transform embeddings for better quantization, followed by an Inverted File Index (IVF) with 65,536 clusters, where the coarse quantizer is accelerated using a Hierarchical Navigable Small World (HNSW) graph. Finally, Product Quantization (PQ) with 32 subquantizers is used to compress vectors for fast and memory-efficient similarity search.
We always feed our models the top $k=2$ retrieved neighbors, both during training and testing.




\section{Impact of Overlap on Perplexity}

As already mentioned, recent work on retrieval-augmented language modeling shows the importance of surface-level overlap between the input chunk and its neighbors in training \RETRO\ \citep{borgeaud2022improving,norlund-etal-2023-generalization,doostmohammadi-etal-2023-surface}.
In our first set of experiments, we want to dive deeper and see how different degrees of overlap affect the training of a \RETRO\ model.
To this end, we artificially bound overlap at predefined thresholds.

\begin{figure}[t]
    \centering
    \hspace*{-3mm} 
    \includegraphics[width=0.5\textwidth]{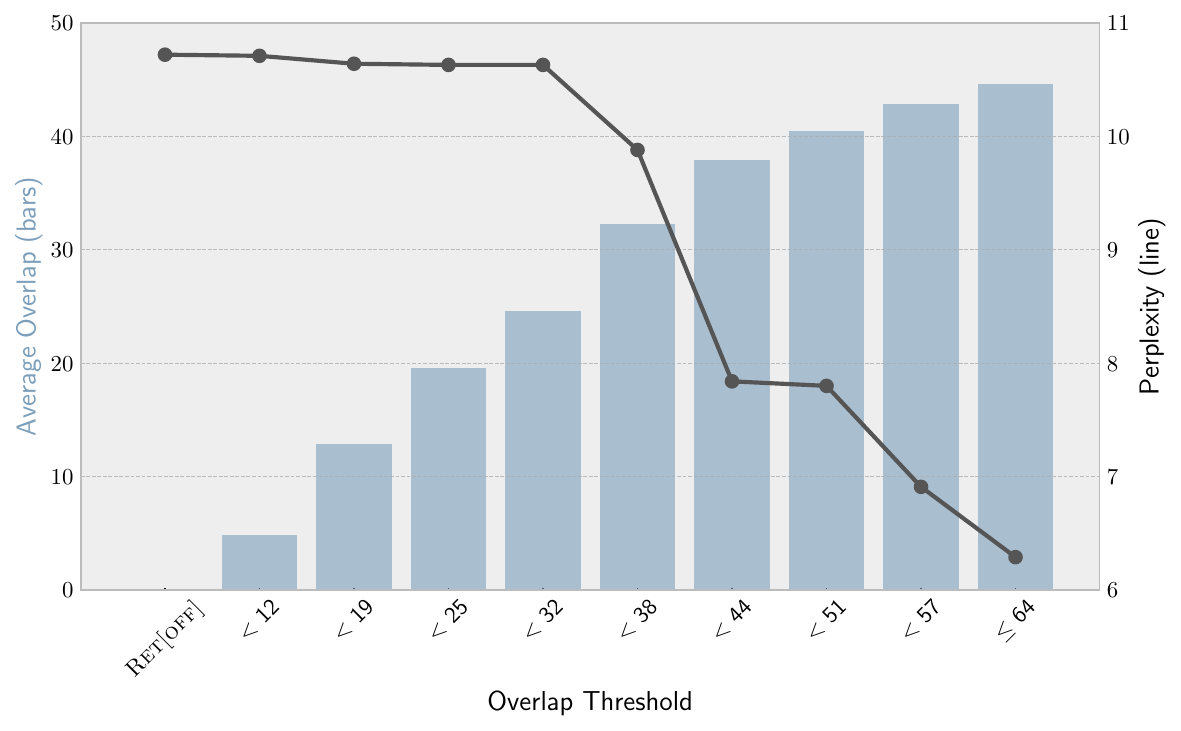}
    \caption{Test perplexity (line) at training step 4,000 and average overlap in terms of number of tokens (bars) for different overlap thresholds during training.}
    \label{fig:plot1}
\end{figure}

\subsection{Overlap Thresholds}

By \emph{overlap}, we mean the number of tokens that are shared between the input chunk and one of its neighbors (including continuations).
We divide the full range of possible overlap values (0--64) into 10 equally-sized (up to rounding) intervals $[\min_i, \max_i]$ ($1 \leq i \leq 10$) and train separate models $M_i$ where we only use neighbors with up to $\max_i$ tokens of overlap (e.g., $<$~25).

During training, for every query chunk, we initially retrieve 20 neighbors and then filter based on the model-specific overlap threshold, prioritizing neighbors with higher overlap.
This approach ensures that we retain only naturally occurring neighbors---a subset of the 20 originally retrieved.
If the number of neighbors with an overlap below the model-specific threshold is less than $k$ (the number of neighbors provided to the model), we substitute the missing neighbors with zero vectors.
We also consider an extreme setting where we provide no neighbors at all, i.e., we replace all neighbors with zeros.
We refer to this setting as \textsc{Ret[off]}.
Note that this differs from a GPT model in that it still uses the additional \RETRO\ parameters.

Note that, during training, while the neighbors are filtered by overlap, the input chunks are the same across all experiments.
At test time, we always use the naturally retrieved neighbors, without any overlap thresholding; the input chunks will vary based on the previously generated tokens.

\subsection{Experimental Setup}
\label{subsec:pretraining_exp}

We train each overlap-thresholded model by \RETRO-fitting the GPT model described in Section~\ref{sec:FrameworkModels} on the Pile \cite{gao2020pile}.
We use the entire training set for retrieval, but only train on a maximum of 10,000 steps with a batch size of 128, which corresponds to approximately 54\% of the full data.
In our training setup, we follow \citet{wang2024instructretro} by using the Adam optimizer \cite{kingma2014adam} with $\beta_1 = 0{.}9$, $\beta_2 = 0{.}98$ and a cosine learning rate decay schedule, starting with a maximum learning rate of 2.5e-4, a minimum of 2.5e-5, and a linear warmup phase spanning the first 5,000 samples.

\begin{figure*}[t]
    \centering
    \hspace*{-0.170\textwidth} 
    \includegraphics[width=1.3\textwidth]{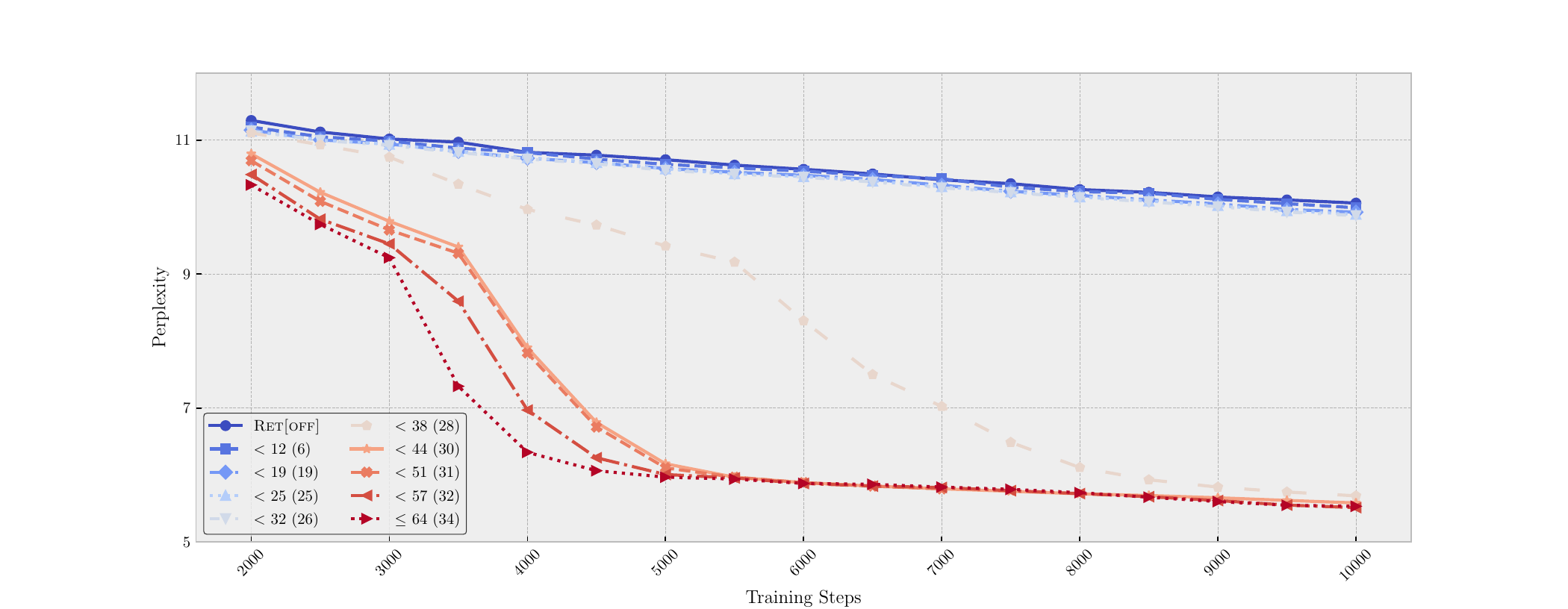}
    \caption{Test perplexity trends for models with different overlap thresholds over training steps. The colors, as well as the line and dot styles, represent the threshold, with colors going from cold to warm as the threshold increases. The legend shows the threshold value, followed by the average overlap for that experiment in parentheses.}
    \label{fig:plot2}
\end{figure*}

\subsection{Results and Analysis}

Figure \ref{fig:plot1} shows the test perplexity and average overlap for each overlap threshold at training step 4,000.
We choose this step here because \citet{borgeaud2022improving} report that it is where \RETRO\ converges; we show the results for other training steps later.
Looking at the plot, we see that, as the threshold increases, the perplexity remains roughly constant up to $<$~32.
However, at the next threshold level, we see a clear drop, and perplexity decreases rapidly as the overlap increases further.
Overall, our results show a strong negative correlation between the test perplexity and the maximal overlap during training.

In Figure \ref{fig:plot2}, we look at perplexity over training steps.
We see that after the threshold $<$~32, all models can reach more or less the same low perplexity given enough time, but the number of steps required for this varies significantly.
The model for threshold $<$~38 takes noticeably longer to converge to a similar perplexity than the models with higher maximal overlap, but the difference among thresholds from $<$~51 to $\leq$~64 is relatively small.
In summary, we find that, while a minimum amount of overlap is needed to ``activate'' a \RETRO\ model, as also discussed in \citet{borgeaud2022improving}, increasing the overlap further leads to faster convergence.
The average overlap for the activated models in this experiment is about 30 tokens.


\section{Overlap and Downstream Tasks}

While we have seen that increased overlap reduces the data requirements for \RETRO-fitting when performance is measured in terms of perplexity, a separate question is whether this benefit carries over to downstream tasks.
To validate this, we apply our overlap-thresholded models to a short-answer generative question answering task.

\begin{figure*}[t]
    \centering
    \hspace*{-0.17\textwidth} 
    \includegraphics[width=1.3\textwidth]{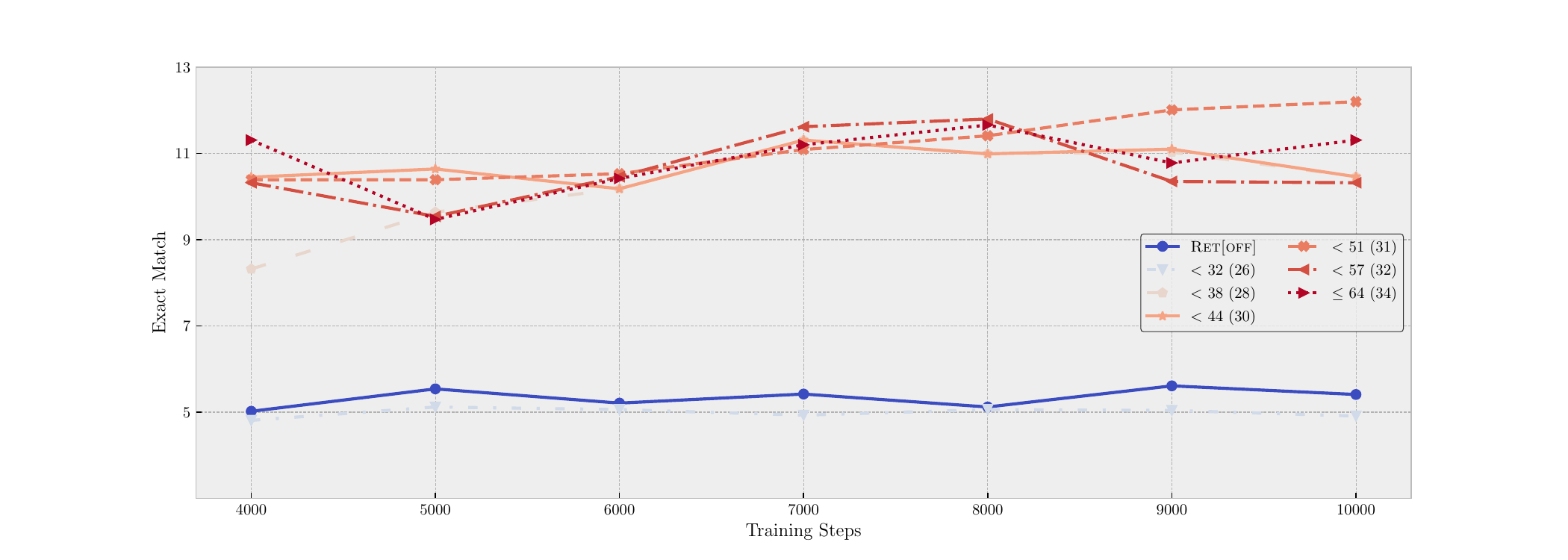}
    \caption{Exact match percentages on the Natural Questions dataset over various training steps and overlap thresholds. The legend shows each threshold value, followed by its corresponding average overlap for that threshold, shown in parentheses.}
    \label{fig:plot3}
\end{figure*}

\begin{figure*}[t]
    \centering
    \hspace*{-0.17\textwidth} 
    \includegraphics[width=1.3\textwidth]{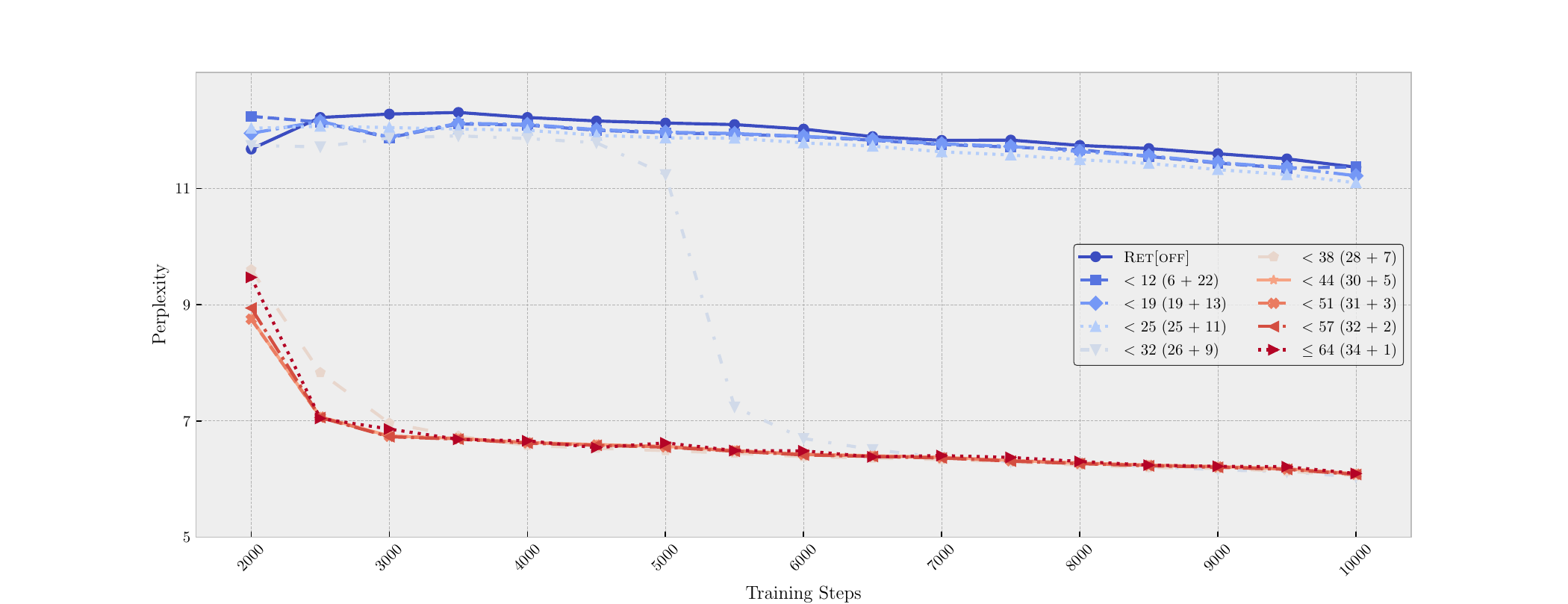}
    \caption{Test perplexity trends of models with varying overlap thresholds over training steps. The legend indicates the threshold value, followed by the average overlap for that threshold and the additional overlap introduced by paraphrases in parentheses.}
    \label{fig:plot4}
\end{figure*}

\subsection{Experimental Setup}
\label{subsec:downstream_exp}

To improve our \RETRO-fitted models' abilities to follow instructions and generate coherent responses, we first instruction-tune them on a blend of open-source datasets provided by \citet{megatronlm_retro_2023}, including Dolly \cite{conover2023free} and Unnatural Instructions \cite{honovich-etal-2023-unnatural}.
We fine-tune the models for 1,000 steps with a batch size of 128.
Following \citet{wang2024instructretro}, we compute the loss only on the answer portion of each question–answer pair and update the weights with a learning rate of 5e-6 and a weight decay of $\lambda = 0{.}01$.
For optimization, we again use Adam with $\beta_1 = 0{.}9$ and $\beta_2 = 0{.}98$.

While \RETRO{} shares its training objective with GPT models, it requires retrieval of nearest neighbors, which many instruction tuning datasets lack.
To avoid using noisy neighbors from the pretraining corpus, we disable the \RETRO\ context encoder using a manually set gate that skips the cross-attention when retrieval is unavailable.
This freezes the encoder parameters and updates only the decoder, simplifying tuning and enabling inference both with and without retrieval.

Following the literature, we evaluate our models on the Natural Questions \cite{kwiatkowski-etal-2019-natural} dataset. We report exact match scores, which means that apart from punctuation marks and whitespace, the model response should exactly match one of the gold answers. Each question in the dataset comes with multiple contexts, which we provide to the model as neighbors with $k=2$. For generation, we use greedy decoding.

\subsection{Results and Analysis}

Figure \ref{fig:plot3} shows the exact match scores of selected models over training steps.
To reduce the number of experiments, we exclude certain models that were not activated in our previous experiments, as all such models had a similar (low) performance.

In general, the results on the downstream task follow a similar pattern to our earlier results on perplexity: unactivated models consistently underperform compared to activated models, which we attribute to their failure to make optimal use of the provided neighbors.
In contrast, activated models benefit from additional training, showing slightly improved exact match scores up to around step 7,000, after which their performance plateaus.
Model $<$~38 initially lags behind---mirroring its perplexity trend---but eventually catches up with the other activated models.
The unactivated models, on the other hand, show a relatively steady (low) performance over different training steps.
In summary, these results validate our earlier perplexity-based findings, showing that perplexity can be used as a predictor of downstream task performance.

\section{A Low-Resource Scenario}
\label{sec:low-resouce}

As shown in Figure \ref{fig:plot2}, activating a \RETRO\ model requires about 4,000 training steps at a batch size of 128, which equals about half a million training samples with highly overlapping neighbors.
This level of data demand, which comes in addition to the data required to pretrain the base model, is often impractical in real-world scenarios---particularly for low-resource languages or specialized domains where such large datasets are unavailable.

To address this limitation, in this section we explore an approach to activating \RETRO\ models by leveraging synthetic data.
Building on our findings regarding the importance of overlap, we propose a methodology that allows us to modulate the strength of the retrieval signal using paraphrased neighbors and thereby control the speed at which a \RETRO\ model gets activated.

\subsection{Experimental Setup}

Our experiments use the same settings and hyperparameters as described in Section \ref{subsec:pretraining_exp} for \RETRO-fitting and Section \ref{subsec:downstream_exp} for instruction tuning and evaluation, with one key difference: we randomly replace one of four chunks in the retrieval-augmented context ($k=2$ neighbors and their continuations) with a paraphrase of the input chunk.

To get these paraphrases, we use the LLaMA 3 8B instruction-tuned model \cite{llama3modelcard} with the prompt provided in Appendix \ref{app:a}.
The prompt is designed to maintain enough surface-level similarity between the input chunk and the paraphrase to get a significant overlap, while at the same time ensuring that the paraphrased chunks are not so similar that the model becomes overly reliant on the newly introduced artificial neighbors.
We then repeat the experiments from the previous sections using the same thresholds as before, but now with the synthesized neighbors added.

\subsection{Results and Analysis}

\paragraph{Impact on perplexity}

Figure \ref{fig:plot4} shows the test perplexity of the overlap-thresholded models over the training steps with paraphrased neighbors.
Paraphrasing, in practice, increases the average overlap per threshold bin.
For consistency, we still retain the original threshold labels and report the new average overlaps as a sum of the old average threshold and the increase introduced through paraphrasing.

Compared to the results before the intervention (Figure \ref{fig:plot2}), models that we previously classified as activated exhibit even faster activation with synthetic neighbors added.
Their convergence curves are also steeper and approach their minimum around step 3,000, compared to step 5,000 in the previous setting. This improvement corresponds to roughly 40\% less data to reach optimal performance.
Model $<$~32, which was not previously activated, eventually reaches the same levels of perplexity as the models with activated retrieval, although it takes approximately 5,500 steps for this to occur.
The price we pay for the faster convergence is higher overall perplexity: the lowest perplexity achieved by the models is now 6.1, compared to 5.6 previously.
A plausible explanation for this observation is that paraphrases introduce noise and decrease data variability.
The detrimental effect on perplexity is more pronounced in models where retrieval is not activated; these do not reach a perplexity below 11, compared to 10 in the setting without synthetic neighbors.

Looking at the average overlap statistics, we see that values increase substantially at lower thresholds, where the paraphrases add a lot of new overlap, while the effect on the higher-threshold models is significantly smaller.
At the same time, convergence happens more quickly for these models regardless, which suggests that factors beyond simple overlap contribute to activating the model's retrieval weights.

\paragraph{Impact on downstream tasks}

Finally, we turn to the downstream results on the Natural Questions dataset, reported in Table \ref{tab:the_only_table}.
We observe a similar trend as in the setup without paraphrases (Figure~\ref{fig:plot3}).
To reduce the number of experiments, we focus on the most relevant thresholds---those near the activation point and the extreme case of $\leq$~64, to assess whether the addition of paraphrases negatively impacts the model under extreme conditions.
The performance approximately doubles once retrieval is activated and remains relatively stable across different training steps.
Overall, while there are some minor fluctuations, the models trained with paraphrased neighbors perform on par with models trained without.
This result demonstrates that, while perplexity is affected negatively, modulating the overlap signal through paraphrasing can lead to faster activation of the augmented retrieval mechanism without degrading downstream performance.

\begin{table}
\centering
\small
\begin{tabular}{lcccc}
\toprule
Thresholds & \multicolumn{4}{c}{\shortstack{Training Steps}} \\
\cmidrule{2-5}
& 4000 & 5000 & 6000 & 7000 \\
\midrule
$<$ 25 \hspace{5pt} \adjustbox{valign=m}{\includegraphics[height=10pt]{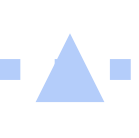}} & 5.5 & 5.4 & 4.9 & 5.3 \\
$<$ 32 \hspace{5pt} \adjustbox{valign=m}{\includegraphics[height=10pt]{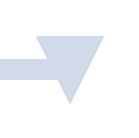}} & 6.1 & 9.8 & 10.8 & 11.1 \\
$<$ 38 \hspace{5pt} \adjustbox{valign=m}{\includegraphics[height=10pt]{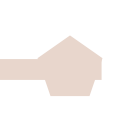}} & 11.3 & 10.1 & 11.2 & 11.4 \\
$<$ 44 \hspace{5pt} \adjustbox{valign=m}{\includegraphics[height=10pt]{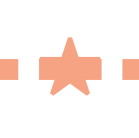}} & 10.3 & 11.3 & 11.5 & 11.0 \\
$\leq$ 64 \hspace{5pt} \adjustbox{valign=m}{\includegraphics[height=10pt]{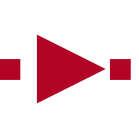}} & 9.8 & 11.4 & 10.5 & 10.7 \\
\bottomrule
\end{tabular}
\caption{Exact match percentages for selected models on the Natural Questions dataset over different training steps. Threshold names and markers match those used in previous plots for ease of comparison.}
\label{tab:the_only_table}
\end{table}

\section{Final Remarks and Discussions}
In this paper, we used a pretrained model, which likely contributed to more stable and robust results and enabled us to conduct more extensive experiments. Moreover, \RETRO-fitting presents a practical and accessible method for users to integrate retrieval mechanisms into language models, given that these models are already pretrained and training from scratch is costly and resource-intensive. However, it remains unclear whether these results would extend to pretraining a \RETRO{} model from random initialization.

Our experiments were conducted using a single model size. While \RETRO{} tends to show greater benefits on smaller models, since they are more constrained in their capacity and must rely more heavily on retrieved neighbors, studying how model size interacts with retrieval-based methods could provide a more comprehensive understanding of their scalability and efficiency.

In our experiments, we only use paraphrasing of the input. However, in low-resource settings, this approach may be limited, as high-quality paraphrasing models might not be available, or off-the-shelf LLMs may struggle to produce fluent paraphrases in the target language. Therefore, other methods, such as back-translation \cite{sugiyama-yoshinaga-2019-data} or synonym substitution \cite{jungiewicz2019towards}, remain to be explored to determine whether they can similarly reduce perplexity without breaking the model.

Although we significantly increased the overlap between neighbors and the input, we were unable to break the model, suggesting that it has a surprisingly high tolerance for redundancy or dependence on neighboring context. However, it is plausible that beyond a certain point, excessive overlap could lead the model to become overly reliant on its neighbors. This dependence may, in turn, cause performance to degrade when such neighbors are absent or differ at test time.

Our results indicate that unigram overlap between the input and its neighbors serves as a useful heuristic and a simple proxy for understanding what drives the model to attend to the neighbors. However, as shown in the results in Section \ref{sec:low-resouce}, it is clearly not the full story. Other factors, such as the stronger signal provided by synthetic data compared to natural language \cite{edunov-etal-2018-understanding}, or variations in word order \cite{norlund-etal-2023-generalization}, may also influence the model’s behavior.

\section{Conclusions and Future Work}

Retrieval-augmented language models have been shown to significantly reduce test-time perplexity, despite their much smaller size compared to standard language models.
Prior work has identified the primary driver of this improvement to be the overlap between the input text and its retrieved neighbors during both training and testing \cite{borgeaud2022improving}. 
Initially, this has no effect, but beyond a certain point, the overlap becomes strong enough to tip the model towards using the retrieved neighbors.

We extend this analysis to different time steps, revealing that more overlap accelerates the model's learning and makes it more likely to attend to retrieval. This suggests that a strong overlap signal between the neighbors and the input chunk is crucial for efficient learning. Additionally, we run experiments on a downstream question-answering task to show that these effects extend beyond just perplexity.

We further extend this by replacing one neighbor with a paraphrase of the input chunk to ensure the model always has a relevant, highly overlapping neighbor. This approach significantly enhances the model's data efficiency, reducing training time by approximately 40\%. We then evaluate the models on a question-answering task to demonstrate that this type of training does not negatively impact model performance. While it is conceivable that further increasing overlap could eventually harm downstream performance or perplexity, we have not observed this in our experiments.

Future research could investigate pretraining a \RETRO{} model from scratch to better understand the challenges and the role of overlap without the stabilizing effects of prior training. Extending the analysis to larger model sizes would also be valuable, as it could reveal how retrieval-based methods scale and whether their benefits persist across capacities. 

Alternative augmentation methods, such as back-translation or synonym substitution, should be explored, especially in low-resource settings where paraphrasing is less viable. Additionally, determining the threshold at which increased input-neighbor overlap causes model failure remains an open question. Finally, further analysis is needed to uncover other factors beyond overlap that influence a model’s attention to its retrieved neighbors.

\section*{Limitations}
While our findings offer valuable insights into the role of input-neighbor overlap in retrieval-augmented language models, several limitations remain. First, our experiments rely on a pretrained language model, leaving open the question of how overlap affects models trained from scratch, where learning dynamics may differ. Second, we only evaluated a single model size; larger models may exhibit different behaviors with respect to retrieval dependence and overlap sensitivity. Third, although we increased overlap extensively without observing model degradation, we did not determine the point at which excessive overlap may begin to harm performance. Finally, while overlap is a convenient and intuitive metric, it likely does not capture the full complexity of retrieval utility—factors such as word order or semantic similarity in non-overlapping tokens warrant further investigation.

\section*{Acknowledgments}
This work was partially supported by the Wallenberg AI, Autonomous Systems and Software Program (WASP) funded by the Knut and Alice Wallenberg Foundation. The computations were enabled by the Berzelius resource provided by the Knut and Alice Wallenberg Foundation at the National Supercomputer Center.

\bibliography{anthology,custom}

\begin{thebibliography}{32}
\expandafter\ifx\csname natexlab\endcsname\relax\def\natexlab#1{#1}\fi

\bibitem[{AI@Meta(2024)}]{llama3modelcard}
AI@Meta. 2024.
\newblock \href {https://github.com/meta-llama/llama3/blob/main/MODEL_CARD.md} {Llama 3 model card}.

\bibitem[{Asai et~al.(2020)Asai, Hashimoto, Hajishirzi, Socher, and Xiong}]{Asai2020Learning}
Akari Asai, Kazuma Hashimoto, Hannaneh Hajishirzi, Richard Socher, and Caiming Xiong. 2020.
\newblock \href {https://openreview.net/forum?id=SJgVHkrYDH} {Learning to retrieve reasoning paths over wikipedia graph for question answering}.
\newblock In \emph{International Conference on Learning Representations}.

\bibitem[{Borgeaud et~al.(2022)Borgeaud, Mensch, Hoffmann, Cai, Rutherford, Millican, Van Den~Driessche, Lespiau, Damoc, Clark et~al.}]{borgeaud2022improving}
Sebastian Borgeaud, Arthur Mensch, Jordan Hoffmann, Trevor Cai, Eliza Rutherford, Katie Millican, George~Bm Van Den~Driessche, Jean-Baptiste Lespiau, Bogdan Damoc, Aidan Clark, et~al. 2022.
\newblock Improving language models by retrieving from trillions of tokens.
\newblock In \emph{International conference on machine learning}, pages 2206--2240. PMLR.

\bibitem[{Conover et~al.(2023)Conover, Hayes, Mathur, Xie, Wan, Shah, Ghodsi, Wendell, Zaharia, and Xin}]{conover2023free}
M.~Conover, M.~Hayes, A.~Mathur, J.~Xie, J.~Wan, S.~Shah, A.~Ghodsi, P.~Wendell, M.~Zaharia, and R.~Xin. 2023.
\newblock \href {https://databricks.com/} {Free dolly: Introducing the world’s first truly open instruction-tuned llm}.
\newblock Technical report, Databricks.

\bibitem[{Doostmohammadi et~al.(2023)Doostmohammadi, Norlund, Kuhlmann, and Johansson}]{doostmohammadi-etal-2023-surface}
Ehsan Doostmohammadi, Tobias Norlund, Marco Kuhlmann, and Richard Johansson. 2023.
\newblock \href {https://doi.org/10.18653/v1/2023.acl-short.45} {Surface-based retrieval reduces perplexity of retrieval-augmented language models}.
\newblock In \emph{Proceedings of the 61st Annual Meeting of the Association for Computational Linguistics (Volume 2: Short Papers)}, pages 521--529, Toronto, Canada. Association for Computational Linguistics.

\bibitem[{Edunov et~al.(2018)Edunov, Ott, Auli, and Grangier}]{edunov-etal-2018-understanding}
Sergey Edunov, Myle Ott, Michael Auli, and David Grangier. 2018.
\newblock \href {https://doi.org/10.18653/v1/D18-1045} {Understanding back-translation at scale}.
\newblock In \emph{Proceedings of the 2018 Conference on Empirical Methods in Natural Language Processing}, pages 489--500, Brussels, Belgium. Association for Computational Linguistics.

\bibitem[{Gao et~al.(2020)Gao, Biderman, Black, Golding, Hoppe, Foster, Phang, He, Thite, Nabeshima et~al.}]{gao2020pile}
Leo Gao, Stella Biderman, Sid Black, Laurence Golding, Travis Hoppe, Charles Foster, Jason Phang, Horace He, Anish Thite, Noa Nabeshima, et~al. 2020.
\newblock The pile: An 800gb dataset of diverse text for language modeling.
\newblock \emph{arXiv preprint arXiv:2101.00027}.

\bibitem[{Guu et~al.(2020)Guu, Lee, Tung, Pasupat, and Chang}]{10.5555/3524938.3525306}
Kelvin Guu, Kenton Lee, Zora Tung, Panupong Pasupat, and Ming-Wei Chang. 2020.
\newblock Realm: retrieval-augmented language model pre-training.
\newblock In \emph{Proceedings of the 37th International Conference on Machine Learning}, ICML'20. JMLR.org.

\bibitem[{Honovich et~al.(2023)Honovich, Scialom, Levy, and Schick}]{honovich-etal-2023-unnatural}
Or~Honovich, Thomas Scialom, Omer Levy, and Timo Schick. 2023.
\newblock \href {https://doi.org/10.18653/v1/2023.acl-long.806} {Unnatural instructions: Tuning language models with (almost) no human labor}.
\newblock In \emph{Proceedings of the 61st Annual Meeting of the Association for Computational Linguistics (Volume 1: Long Papers)}, pages 14409--14428, Toronto, Canada. Association for Computational Linguistics.

\bibitem[{Izacard et~al.(2023)Izacard, Lewis, Lomeli, Hosseini, Petroni, Schick, Dwivedi-Yu, Joulin, Riedel, and Grave}]{izacard2023atlas}
Gautier Izacard, Patrick Lewis, Maria Lomeli, Lucas Hosseini, Fabio Petroni, Timo Schick, Jane Dwivedi-Yu, Armand Joulin, Sebastian Riedel, and Edouard Grave. 2023.
\newblock Atlas: Few-shot learning with retrieval augmented language models.
\newblock \emph{Journal of Machine Learning Research}, 24(251):1--43.

\bibitem[{Johnson et~al.(2019)Johnson, Douze, and J{\'e}gou}]{johnson2019billion}
Jeff Johnson, Matthijs Douze, and Herv{\'e} J{\'e}gou. 2019.
\newblock Billion-scale similarity search with {GPUs}.
\newblock \emph{IEEE Transactions on Big Data}, 7(3):535--547.

\bibitem[{Jungiewicz and Smywi{\'n}ski-Pohl(2019)}]{jungiewicz2019towards}
Micha{\l} Jungiewicz and Aleksander Smywi{\'n}ski-Pohl. 2019.
\newblock Towards textual data augmentation for neural networks: synonyms and maximum loss.
\newblock \emph{Computer Science}, 20.

\bibitem[{Karpukhin et~al.(2020)Karpukhin, Oguz, Min, Lewis, Wu, Edunov, Chen, and Yih}]{karpukhin-etal-2020-dense}
Vladimir Karpukhin, Barlas Oguz, Sewon Min, Patrick Lewis, Ledell Wu, Sergey Edunov, Danqi Chen, and Wen-tau Yih. 2020.
\newblock \href {https://doi.org/10.18653/v1/2020.emnlp-main.550} {Dense passage retrieval for open-domain question answering}.
\newblock In \emph{Proceedings of the 2020 Conference on Empirical Methods in Natural Language Processing (EMNLP)}, pages 6769--6781, Online. Association for Computational Linguistics.

\bibitem[{Khandelwal et~al.(2020)Khandelwal, Levy, Jurafsky, Zettlemoyer, and Lewis}]{Khandelwal2020Generalization}
Urvashi Khandelwal, Omer Levy, Dan Jurafsky, Luke Zettlemoyer, and Mike Lewis. 2020.
\newblock \href {https://openreview.net/forum?id=HklBjCEKvH} {Generalization through memorization: Nearest neighbor language models}.
\newblock In \emph{International Conference on Learning Representations}.

\bibitem[{Kingma and Ba(2014)}]{kingma2014adam}
Diederik~P Kingma and Jimmy Ba. 2014.
\newblock Adam: A method for stochastic optimization.
\newblock \emph{arXiv preprint arXiv:1412.6980}.

\bibitem[{Kwiatkowski et~al.(2019)Kwiatkowski, Palomaki, Redfield, Collins, Parikh, Alberti, Epstein, Polosukhin, Devlin, Lee, Toutanova, Jones, Kelcey, Chang, Dai, Uszkoreit, Le, and Petrov}]{kwiatkowski-etal-2019-natural}
Tom Kwiatkowski, Jennimaria Palomaki, Olivia Redfield, Michael Collins, Ankur Parikh, Chris Alberti, Danielle Epstein, Illia Polosukhin, Jacob Devlin, Kenton Lee, Kristina Toutanova, Llion Jones, Matthew Kelcey, Ming-Wei Chang, Andrew~M. Dai, Jakob Uszkoreit, Quoc Le, and Slav Petrov. 2019.
\newblock \href {https://doi.org/10.1162/tacl_a_00276} {Natural questions: A benchmark for question answering research}.
\newblock \emph{Transactions of the Association for Computational Linguistics}, 7:452--466.

\bibitem[{{Megatron}(2023)}]{megatronlm_retro_2023}
{Megatron}. 2023.
\newblock {Nvidia Megatron-LM: tools/retro (commit 47e3bd3)}.
\newblock \url{https://github.com/NVIDIA/Megatron-LM/tree/47e3bd3047aafbae361e1699d1d8785d678732ca/tools/retro}.
\newblock Accessed: 2025-05-05.

\bibitem[{Norlund et~al.(2023)Norlund, Doostmohammadi, Johansson, and Kuhlmann}]{norlund-etal-2023-generalization}
Tobias Norlund, Ehsan Doostmohammadi, Richard Johansson, and Marco Kuhlmann. 2023.
\newblock \href {https://doi.org/10.18653/v1/2023.findings-eacl.109} {On the generalization ability of retrieval-enhanced transformers}.
\newblock In \emph{Findings of the Association for Computational Linguistics: EACL 2023}, pages 1485--1493, Dubrovnik, Croatia. Association for Computational Linguistics.

\bibitem[{Radford et~al.(2019)Radford, Wu, Child, Luan, Amodei, Sutskever et~al.}]{radford2019language}
Alec Radford, Jeffrey Wu, Rewon Child, David Luan, Dario Amodei, Ilya Sutskever, et~al. 2019.
\newblock Language models are unsupervised multitask learners.

\bibitem[{Reimers and Gurevych(2019)}]{reimers-2019-sentence-bert}
Nils Reimers and Iryna Gurevych. 2019.
\newblock \href {http://arxiv.org/abs/1908.10084} {Sentence-bert: Sentence embeddings using siamese bert-networks}.
\newblock In \emph{Proceedings of the 2019 Conference on Empirical Methods in Natural Language Processing}. Association for Computational Linguistics.

\bibitem[{Samuel et~al.(2024)Samuel, Charpentier, and Wold}]{samuel-etal-2024-room}
David Samuel, Lucas Charpentier, and Sondre Wold. 2024.
\newblock \href {https://doi.org/10.18653/v1/2024.naacl-short.26} {More room for language: Investigating the effect of retrieval on language models}.
\newblock In \emph{Proceedings of the 2024 Conference of the North American Chapter of the Association for Computational Linguistics: Human Language Technologies (Volume 2: Short Papers)}, pages 282--305, Mexico City, Mexico. Association for Computational Linguistics.

\bibitem[{Shi et~al.(2024{\natexlab{a}})Shi, Min, Lomeli, Zhou, Li, Lin, Smith, Zettlemoyer, tau Yih, and Lewis}]{shi2024incontext}
Weijia Shi, Sewon Min, Maria Lomeli, Chunting Zhou, Margaret Li, Xi~Victoria Lin, Noah~A. Smith, Luke Zettlemoyer, Wen tau Yih, and Mike Lewis. 2024{\natexlab{a}}.
\newblock \href {https://openreview.net/forum?id=LXVswInHOo} {In-context pretraining: Language modeling beyond document boundaries}.
\newblock In \emph{The Twelfth International Conference on Learning Representations}.

\bibitem[{Shi et~al.(2024{\natexlab{b}})Shi, Tan, Wu, Zhong, Zhou, and Liu}]{10.1145/3627673.3679722}
Yucheng Shi, Qiaoyu Tan, Xuansheng Wu, Shaochen Zhong, Kaixiong Zhou, and Ninghao Liu. 2024{\natexlab{b}}.
\newblock \href {https://doi.org/10.1145/3627673.3679722} {Retrieval-enhanced knowledge editing in language models for multi-hop question answering}.
\newblock In \emph{Proceedings of the 33rd ACM International Conference on Information and Knowledge Management}, CIKM '24, page 2056–2066, New York, NY, USA. Association for Computing Machinery.

\bibitem[{Shoeybi et~al.(2019)Shoeybi, Patwary, Puri, LeGresley, Casper, and Catanzaro}]{shoeybi2019megatron}
Mohammad Shoeybi, Mostofa Patwary, Raul Puri, Patrick LeGresley, Jared Casper, and Bryan Catanzaro. 2019.
\newblock Megatron-lm: Training multi-billion parameter language models using model parallelism.
\newblock \emph{arXiv preprint arXiv:1909.08053}.

\bibitem[{Sugiyama and Yoshinaga(2019)}]{sugiyama-yoshinaga-2019-data}
Amane Sugiyama and Naoki Yoshinaga. 2019.
\newblock \href {https://doi.org/10.18653/v1/D19-6504} {Data augmentation using back-translation for context-aware neural machine translation}.
\newblock In \emph{Proceedings of the Fourth Workshop on Discourse in Machine Translation (DiscoMT 2019)}, pages 35--44, Hong Kong, China. Association for Computational Linguistics.

\bibitem[{Wang et~al.(2024)Wang, Ping, McAfee, Xu, Li, Shoeybi, and Catanzaro}]{wang2024instructretro}
Boxin Wang, Wei Ping, Lawrence McAfee, Peng Xu, Bo~Li, Mohammad Shoeybi, and Bryan Catanzaro. 2024.
\newblock Instructretro: instruction tuning post retrieval-augmented pretraining.
\newblock In \emph{Proceedings of the 41st International Conference on Machine Learning}, pages 51255--51272.

\bibitem[{Wang et~al.(2023{\natexlab{a}})Wang, Ping, Xu, McAfee, Liu, Shoeybi, Dong, Kuchaiev, Li, Xiao, Anandkumar, and Catanzaro}]{wang2023shall}
Boxin Wang, Wei Ping, Peng Xu, Lawrence McAfee, Zihan Liu, Mohammad Shoeybi, Yi~Dong, Oleksii Kuchaiev, Bo~Li, Chaowei Xiao, Anima Anandkumar, and Bryan Catanzaro. 2023{\natexlab{a}}.
\newblock Shall we pretrain autoregressive language models with retrieval? a comprehensive study.

\bibitem[{Wang et~al.(2023{\natexlab{b}})Wang, Liu, Yue, Tang, Zhang, Jiayang, Yao, Gao, Hu, Qi et~al.}]{wang2023survey}
Cunxiang Wang, Xiaoze Liu, Yuanhao Yue, Xiangru Tang, Tianhang Zhang, Cheng Jiayang, Yunzhi Yao, Wenyang Gao, Xuming Hu, Zehan Qi, et~al. 2023{\natexlab{b}}.
\newblock Survey on factuality in large language models: Knowledge, retrieval and domain-specificity.
\newblock \emph{arXiv preprint arXiv:2310.07521}.

\bibitem[{Wang et~al.(2020)Wang, Wei, Dong, Bao, Yang, and Zhou}]{wang2020minilm}
Wenhui Wang, Furu Wei, Li~Dong, Hangbo Bao, Nan Yang, and Ming Zhou. 2020.
\newblock \href {http://arxiv.org/abs/2002.10957} {Minilm: Deep self-attention distillation for task-agnostic compression of pre-trained transformers}.

\bibitem[{Xu et~al.(2023)Xu, Alon, and Neubig}]{10.5555/3618408.3620004}
Frank~F. Xu, Uri Alon, and Graham Neubig. 2023.
\newblock Why do nearest neighbor language models work?
\newblock In \emph{Proceedings of the 40th International Conference on Machine Learning}, ICML'23. JMLR.org.

\bibitem[{Yogatama et~al.(2021{\natexlab{a}})Yogatama, de~Masson~d{'}Autume, and Kong}]{yogatama-etal-2021-adaptive}
Dani Yogatama, Cyprien de~Masson~d{'}Autume, and Lingpeng Kong. 2021{\natexlab{a}}.
\newblock \href {https://doi.org/10.1162/tacl_a_00371} {Adaptive semiparametric language models}.
\newblock \emph{Transactions of the Association for Computational Linguistics}, 9:362--373.

\bibitem[{Yogatama et~al.(2021{\natexlab{b}})Yogatama, de~Masson~d’Autume, and Kong}]{10.1162/tacl_a_00371}
Dani Yogatama, Cyprien de~Masson~d’Autume, and Lingpeng Kong. 2021{\natexlab{b}}.
\newblock \href {https://doi.org/10.1162/tacl_a_00371} {Adaptive semiparametric language models}.
\newblock \emph{Transactions of the Association for Computational Linguistics}, 9:362--373.

\end{thebibliography}
\bibliographystyle{acl_natbib}

\appendix

\section{Paraphrasing Prompt Template}
\label{app:a}

\noindent We use the following prompt in our experiments for paraphrasing text chunks:

\begin{quote}
\small
\texttt{Paraphrase the following text.\\
- Keep the meaning and the overall structure the same, but you can change the words.\\
- The paraphrase should have the same length as the input.\\
- Strictly keep the order of the information intact.\\
- DO NOT add and DO NOT remove any information.\\
- Only generate a JSON like \{\{"paraphrase": THE\_PARAPHRASED\_TEXT\_HERE\}\} and nothing more:\textbackslash n\textbackslash n\{chunk\}}
\end{quote}



\end{document}